\documentclass[preprint,times,sort&compress]{elsarticle}

\pdfoutput=1

\journal{arXiv}

\usepackage{fixltx2e}

\usepackage{url}
\usepackage[breaklinks, hidelinks]{hyperref}

\usepackage{graphicx}

\usepackage{algorithmic}
\usepackage{algorithm}

\usepackage{amsmath}
\usepackage{mathtools}

\usepackage{multirow}
\usepackage{booktabs}
\usepackage{makecell}

\usepackage[format=hang, labelformat=parens]{subcaption}
\usepackage{siunitx}

\usepackage[T1]{fontenc}

\begin{document}

\begin{frontmatter}

\title{The observer-assisted method for adjusting hyper-parameters in deep learning algorithms}

\author[aghaddress]{Maciej Wielgosz}
\ead{wielgosz@agh.edu.pl}

\address[aghaddress]{AGH University of Science and Technology, Krak\'ow, Poland}

\begin{abstract}
This paper presents a concept of a novel method for adjusting hyper-parameters in Deep Learning (DL) algorithms. An external agent-observer monitors a performance of a selected Deep Learning algorithm. The observer learns to model the DL algorithm using a series of random experiments. Consequently, it may be used for predicting a response of the DL algorithm in terms of a selected quality measurement to a set of hyper-parameters. This allows to construct an ensemble composed of a series of evaluators which constitute an observer-assisted architecture. The architecture may be used to gradually iterate towards to the best achievable quality score in tiny steps governed by a unit of progress. The algorithm is stopped when the maximum number of steps is reached or no further progress is made.

\end{abstract}

\begin{keyword}
Deep Learning, Hyper-Parameters, Optimization
\end{keyword}

\end{frontmatter}


\section{Introduction}
\label{section:introduction}

\begin{figure}
\centering
\includegraphics[width=0.45\textwidth]{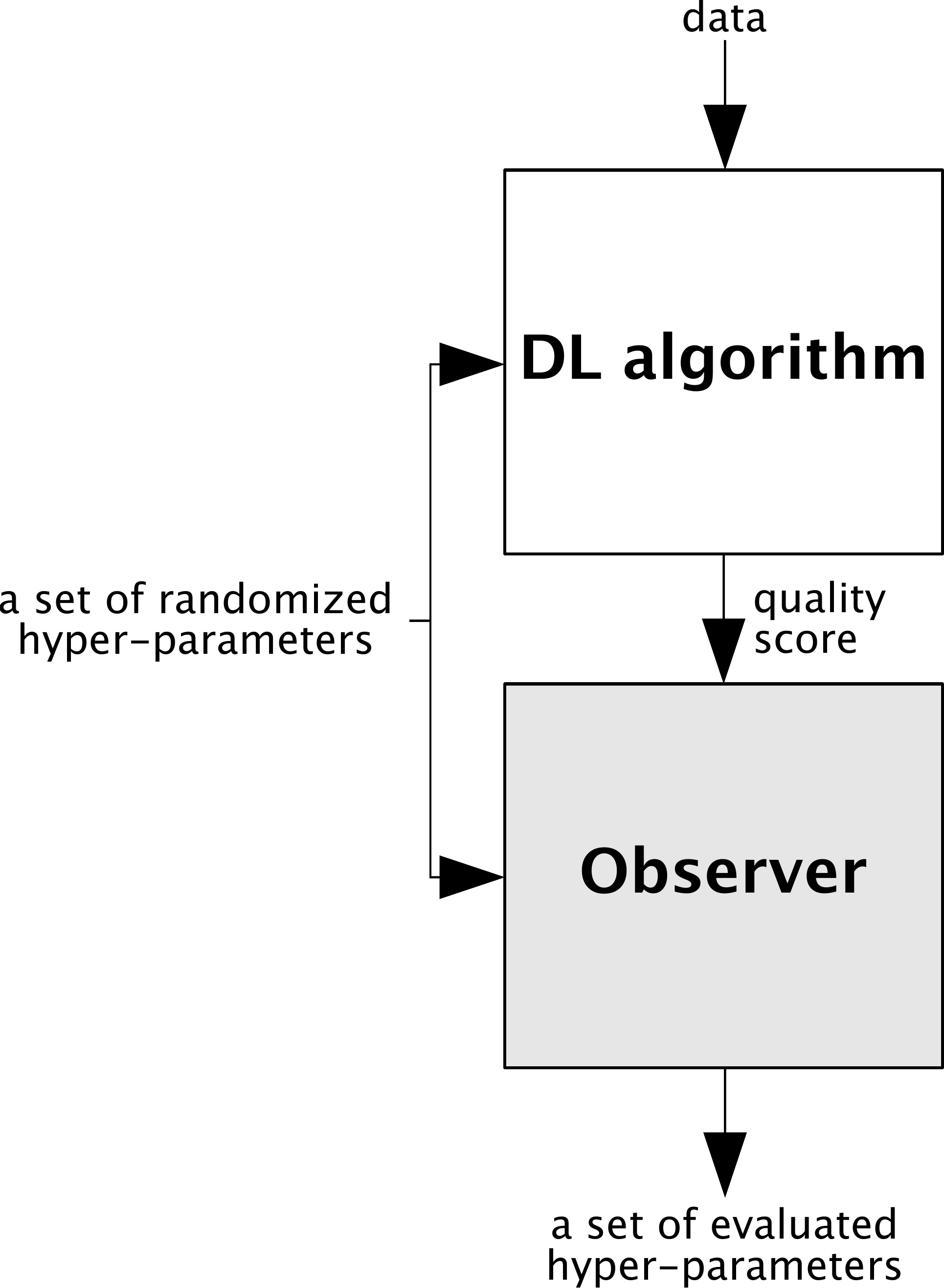}
\caption{Block diagram of the observer-assisted adjusting of hyper-parameters.}
\label{fig:observer-idea}
\end{figure}

Hyper-parameters adjusting is a challenging task which was addressed in many papers \cite{bergstra2012random, talathi2015hyper, snoek2012practical, bengio2012practical}. It is important, because virtually all the currently used algorithms feature macro parameters, which shape their final architecture. This in turn has a direct impact on a performance of solutions based on those algorithms. Unfortunately, despite the fact that deep learning algorithms have been around for a long time, there are no well-established procedures for hyper-parameters tuning, such as back-propagation for a model training \cite{Li2012Brief}. Instead, a set of custom techniques, such as grid, random and heuristic search \cite{donoghue2015framework, young2015optimizing}, have been developed and used by most of DL systems designers.

In the author's view a process of adjusting hyper-parameters should account for both data and the algorithm. This in turn requires an external agent denoted herein as `observer'. Such an observer learns how a given set of hyper-parameters affects a performance of a deep learning algorithm in terms of a chosen quality measurement such as F1 score or accuracy (Fig.~\ref{fig:observer-idea}).

The basic idea is to offset a learning process from a complex AI algorithm which is hard to control to a simpler one with easily adjustable set of hyper-parameters \cite{bengio2012practical}. For instance choosing a set of hyper-parameters for Hierarchical Temporal Memory (HTM) \cite{ahmad2016how, ahmad2015properties, wielgosz2016using} is very time consuming and demanding process. It can be replaced by using feed-forward neural network to model HTM and predict the right values of hyper-parameters such as number of cells, columns and synapses. 
\section{Algorithm}
\label{section:algorithm}

The observer learns a response of a deep learning algorithm for different sets of hyper-parameters in terms of a selected quality measurement score (Fig.~\ref{fig:observer-idea}). Based on this information, it is able to reason about the best set of the hyper-parameters. However, in order to be able to learn a relationship between the hyper-parameters and the performance, a range of experiments with a random set of the parameters must be conducted. The more experiments are done, the more reliable predictions can be made by the observer. It is worth keeping in mind that the observer models the deep learning algorithm. Therefore, a quality of a model used as the observer has a substantial impact on the hyper-parameters being adjusted. 

\begin{table}
\caption{Basic algorithm notation}\label{tab:notation:basic}
\centering
\begin{tabular}{cl}
\toprule
$n$ & number of hyper-parameters \\ \midrule
$\mathit{hp}_0, \mathit{hp}_1, \dots, \mathit{hp}_{n-1}$  & input hyper-parameters \\ \midrule
$\mathit{hp\_eval}_0, \mathit{hp\_eval}_1, \dots, \mathit{hp\_eval}_{n-1}$  & evaluated hyper-parameters \\ \midrule
$\mathit{hp\_best}_0, \mathit{hp\_best}_1, \dots, \mathit{hp\_best}_{n-1}$  & \makecell[l]{evaluated hyper-parameters yielding best \\ quality score during algorithm lifetime} \\ \midrule
$\mathit{q\_ex}$ & expected quality score; $\mathit{q\_ex} \in \left[ 0, 1 \right]$ \\ \midrule
$\mathit{q\_eval}_0, \mathit{q\_eval}_1, \dots, \mathit{q\_eval}_{n-1}$ & evaluated quality scores; $\mathit{q\_eval}_\mathit{idx} \in \left[ 0, 1 \right]$ \\ \midrule
$\mathit{q\_best}$ & \makecell[l]{best evaluated quality score during algorithm \\ lifetime; $\mathit{q\_best} \in \left[ 0, 1 \right]$} \\ \midrule
$\mathrm{map}_0, \mathrm{map}_1, \dots, \mathrm{map}_{n-1}$ & \makecell[l]{ mapping functions evaluating \\ hyper-parameters $0 \dots n-1$} \\ \midrule
$\mathrm{map\_q}$ & mapping function evaluating quality score \\ \midrule
$\mathit{iterations}$ & iterations counter \\ \midrule
$\mathit{max\_iterations}$ & \makecell[l]{number of iterations after which the training \\ is stopped} \\ \midrule
$\mathit{min\_contribution}$ & \makecell[l]{minimum expected change in the quality \\ score between iterations} \\ \midrule
$\mathit{idle}$ & \makecell[l]{number of consecutive iterations in which \\ $\mathrm{max}(\mathit{q\_eval}) - \mathit{q\_best} < \mathit{min\_contribution}$} \\ \midrule
$\mathit{max\_idle}$ & \makecell[l]{number of idle iterations after which \\ the training is stopped} \\
\bottomrule
\end{tabular}
\end{table}

\begin{figure}
\centering
\includegraphics[width=0.85\textwidth]{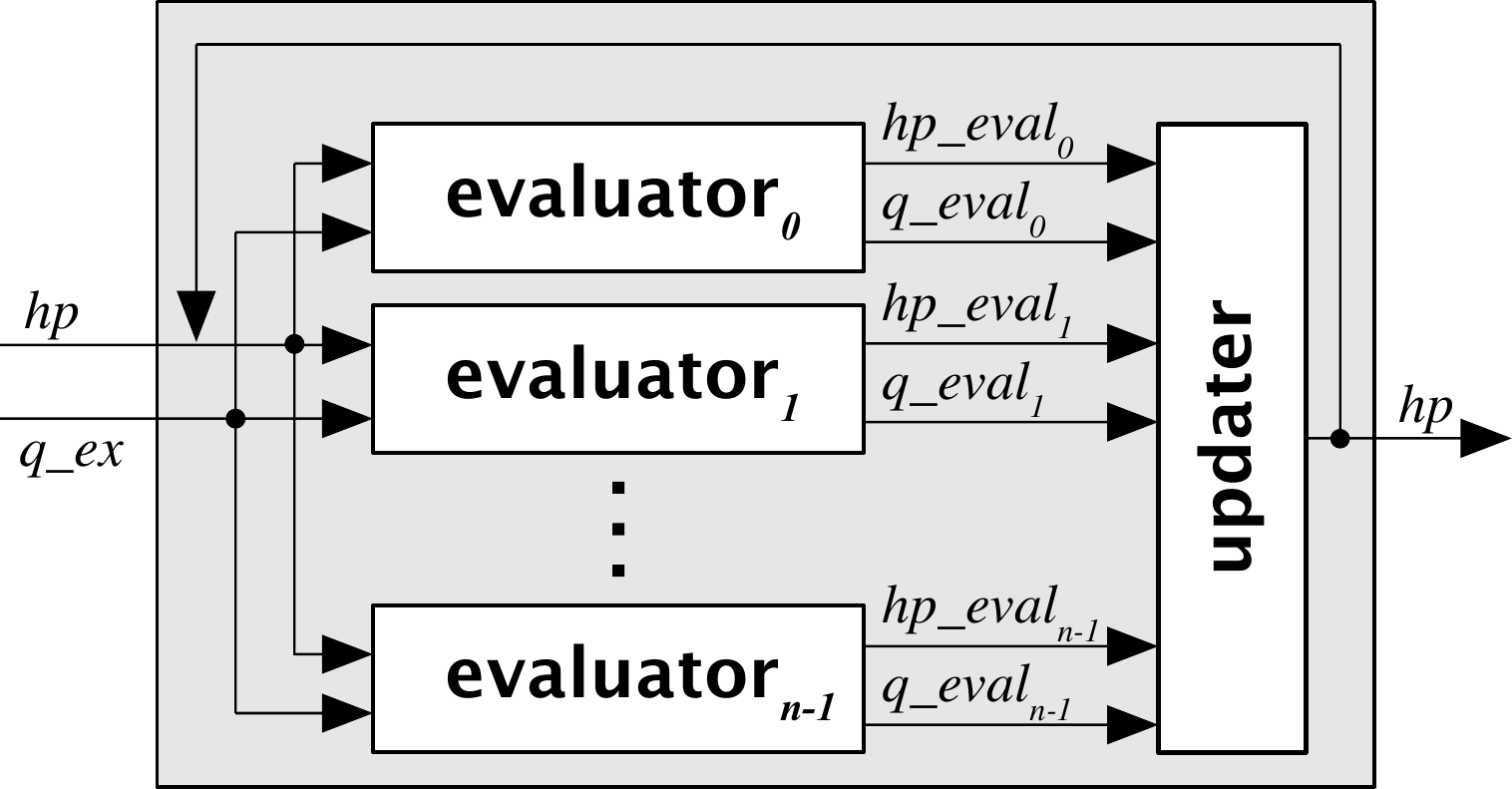}
\caption{Structure of the observer for $n$ hyper-parameters tuning.}
\label{fig:observer-basic}
\end{figure}

\begin{figure}
\centering
\includegraphics[width=0.75\textwidth]{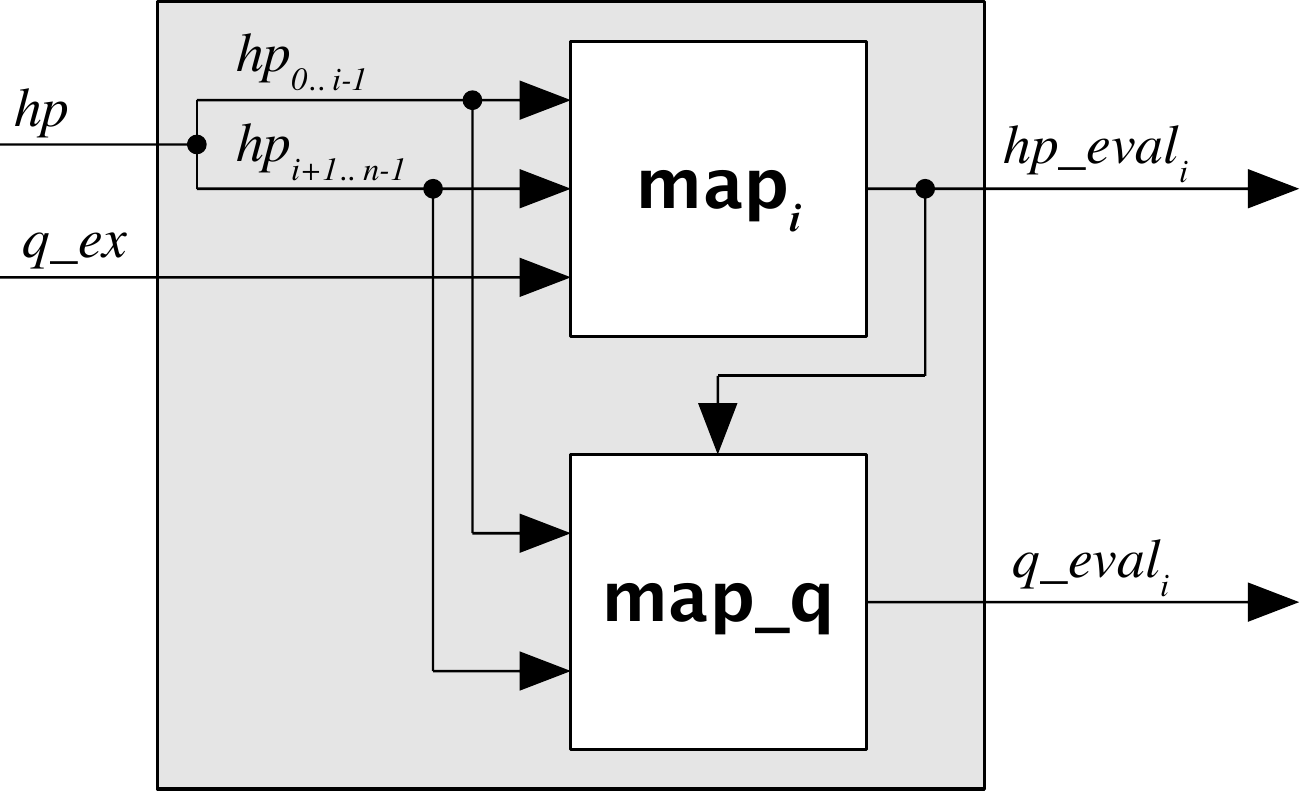}
\caption{Structure of the single evaluator module for $n$ hyper-parameters tuning.}
\label{fig:evaluator}
\end{figure}

A key component of the observer algorithm is presented in Fig.~\ref{fig:observer-basic}, where a series of evaluators are shown. Each of the evaluators provides information regarding a value of a selected hyper-parameter and quality score. Taking the evaluators outputs into account, a decision about which hyper-parameter is updated in the next step is made by the updater module. Each evaluator consist of two mappers (Fig.~\ref{fig:evaluator}). The first mapper, unique for each evaluator, is responsible for hyper-parameter value evaluation. The second one is used to predict quality score for a given set of hyper-parameters. In general, mappers may be implemented as any kind of an algorithm such as CNN, linear or logistic regression \cite{zeiler2014visualizing}. It is worth keeping in mind that a regression algorithm as such is not as important as its prediction quality. However, poor prediction does not necessarily mean lack of the algorithm convergence.

\subsection{Basic algorithm}
\label{subsection:algorithm:basic}

The observer-assisted hyper-parameters adjusting algorithm presented in Alg.~\ref{alg:basic} assumes predefined values of $\mathit{max\_iterations}$, $\mathit{max\_idle}$ and $\mathit{min\_contribution}$, as well as expected (target) quality score $\mathit{q\_ex}$. The author of the paper assumed that the threshold can be arbitrarily chosen, but in practice there are probably very few cases when a designer expects quality score $\mathit{q\_ex} < 1$.

\begin{algorithm}
\caption{Basic hyper-parameters adjusting}
\label{alg:basic}
\begin{algorithmic}[1]
\REQUIRE $\mathit{q\_ex}$
\ENSURE $\mathit{q\_best}, \mathit{hp\_best}$
\STATE $\mathit{hp} \leftarrow$ get\_random\_hyperparams()
\STATE $\mathit{hp\_best} \leftarrow \mathit{hp}$
\STATE $\mathit{iterations} \leftarrow 0$
\STATE $\mathit{idle} \leftarrow 0$
\STATE $\mathit{q\_best} \leftarrow \mathrm{map\_q}(\mathit{hp}_0, \dots, \mathit{hp}_{n-1})$

 \WHILE{$\mathit{q\_best} < \mathit{q\_ex}$ \AND $\mathit{idle} < \mathit{max\_idle}$ \AND $\mathit{iterations} < \mathit{max\_iterations}$}
    \FOR{$i=0$ \TO $n-1$} \label{alg:basic:lst:hp_eval:begin}
        \STATE $\mathit{hp\_eval}_i \leftarrow \mathrm{map}_i(\mathit{hp}_0, \dots, \mathit{hp}_{i-1}, \mathit{hp}_{i+1}, \dots, \mathit{hp}_{n-1}, \mathit{q\_ex})$
    \ENDFOR \label{alg:basic:lst:hp_eval:end}
    \FOR{$i=0$ \TO $n-1$} \label{alg:basic:lst:f1_eval:begin}
        \STATE $\mathit{q\_eval}_i \leftarrow \mathrm{map\_q}(\mathit{hp}_0, \dots, \mathit{hp}_{i-1}, \mathit{hp\_eval}_i, \mathit{hp}_{i+1}, \dots, \mathit{hp}_{n-1})$ 
    \ENDFOR \label{alg:basic:lst:f1_eval:end}
    
    \STATE $\mathit{idx} \leftarrow$ get\_updated\_hyperparam\_index($\mathit{q\_eval}$) \label{alg:basic:lst:get_index}
    
    \STATE $\mathit{hp}_{\mathit{idx}} \leftarrow \mathit{hp\_eval}_{\mathit{idx}}$
    
    \IF{$\mathit{q\_eval}_{\mathit{idx}} - \mathit{q\_best} > \mathit{min\_contribution}$}
        \STATE $\mathit{q\_best} \leftarrow \mathit{q\_eval}_{\mathit{idx}}$
        \STATE $\mathit{hp\_best} \leftarrow \mathit{hp}$
        \STATE $\mathit{idle} \leftarrow 0$
    \ELSE
        \STATE $\mathit{idle} \leftarrow \mathit{idle} + 1$
    \ENDIF
    
    \STATE {$\mathit{iterations} \leftarrow \mathit{iterations} + 1$} 
 \ENDWHILE
\end{algorithmic}
\end{algorithm}

\begin{algorithm}
\caption{Basic updated hyper-parameter selection (see Alg.~\ref{alg:basic} step \ref{alg:basic:lst:get_index})}
\label{alg:get_index:basic}
\begin{algorithmic}[1]
\REQUIRE $\mathit{q\_eval}$
\ENSURE $\mathit{idx}$
\STATE $\mathit{idx} \leftarrow 0$ 
    \FOR{$i=1$ \TO $n-1$}
        \IF{$\mathit{q\_eval}_{\mathit{idx}} < \mathit{q\_eval}_i$}
            \STATE $\mathit{idx} \leftarrow i$
        \ENDIF
    \ENDFOR
\end{algorithmic}
\end{algorithm}

In the first part of the algorithm, hyper-parameters $\mathit{hp}_0, \mathit{hp}_1, \dots, \mathit{hp}_{n-1}$ are initialized with random values from appropriate ranges, iterations counter, idle iterations counter and current quality score value are set to $0$. Iterations counter is used to prevent algorithm from running infinitely. Idle iterations counter is used to stop the adjusting process after a certain time when there is no progress.

The main work of the algorithm is done inside the while loop. In steps \ref{alg:basic:lst:hp_eval:begin} -- \ref{alg:basic:lst:hp_eval:end}, each mapper evaluates a hyper-parameter value based on the current hyper-parameters set and expected quality score. This results in a creation of an array with the proposed values of all hyper-parameters that can potentially yield better results than current set. In steps \ref{alg:basic:lst:f1_eval:begin} -- \ref{alg:basic:lst:f1_eval:end} those propositions are evaluated with predicted quality scores stored in $\mathit{q\_eval}$ array. After that, a single parameter, for which evaluated quality score was the highest, is chosen as a replacement in the current hyper-parameters set (Alg.~\ref{alg:get_index:basic}). If best evaluated quality score is better than previous best one, it is (along with its parameters set) remembered for the future and idle iterations counter is reset. Otherwise, iteration is counted as idle.

\subsection{Algorithm with increasing expected quality score (multi-pass approach)}
\label{subsection:algorithm:multi-pass}

The proposed Alg.~\ref{alg:basic} approaches hyper-parameters adjusting problem in a single pass, i.e. it sets the expected quality score to the highest possible value (e.g. $\mathit{q\_ex} = 1$). Another option is to reach target quality score in multiple passes (shown in Alg.~\ref{alg:multi-pass}), increasing expected quality score a little with each pass. Such a strategy may lead to a faster algorithm convergence and/or prevent it from being unable to leave a local minimum.

\begin{table}
\caption{Multi-pass approach notation (in addition to Tab.~\ref{tab:notation:basic})}\label{tab:notation:multi-pass}
\centering
\begin{tabular}{cl}
\toprule
$\mathit{q\_target}$ & target quality score \\ \midrule
$\mathit{q\_init}$ & initial (minimum) expected quality score \\ \midrule
$\mathit{q\_step}$ & expected quality score increment \\ \midrule
$\mathit{stagnation}$ & \makecell[l]{counter of consecutive passes failing to reach expected \\ quality score} \\ \midrule
$\mathit{max\_stagnation}$ & $\mathit{stagnation}$ counter upper limit \\
\bottomrule
\end{tabular}
\end{table}

\begin{algorithm}
\caption{Multi-pass hyper-parameters adjusting}
\label{alg:multi-pass}
\begin{algorithmic}[1]
\STATE $\mathit{q\_ex} \leftarrow \mathit{q\_init}$
\STATE $\mathit{stagnation} \leftarrow 0$

\WHILE{$\mathit{q\_best} < \mathit{q\_target}$ \AND $\mathit{stagnation} < \mathit{max\_stagnation}$}
    \STATE $\mathit{q\_best},  \mathit{hp\_best} \leftarrow $ run\_basic\_algorithm($\mathit{q\_ex}$)
    \IF{$\mathit{q\_best} < \mathit{q\_ex}$}
        \STATE $\mathit{stagnation} \leftarrow \mathit{stagnation} + 1$
    \ELSE
        \STATE $\mathit{stagnation} \leftarrow 0$
    \ENDIF
    \STATE $\mathit{q\_step} \leftarrow $ get\_q\_step()
    \STATE $\mathit{q\_ex} \leftarrow \mathit{q\_ex} + \mathit{q\_step}$
\ENDWHILE
\end{algorithmic}
\end{algorithm}

Initially, $\mathit{q\_ex} = \mathit{q\_init}$. In the concurrent passes of the algorithm, $\mathit{q\_ex}$ value is increased by $\mathit{q\_step}$. There are two strategies of choosing  $\mathit{q\_step}$: it may be constant and small throughout all the passes or it may be progressively decreased with a rising number of passes. The decision about starting a new pass depends on the algorithm achieving the previously set expected quality score and whether $\mathit{stagnation}$ counter reached its upper limit.

\subsection{Algorithm with a modified updated hyper-parameter selection (cost-based)}
\label{subsection:algorithm:cost}

In basic version of the algorithm, selection of the updated hyper-parameter is done using a simple criteria of evaluated quality score comparison, with hyper-parameter yielding the highest quality score being chosen (see Alg.~\ref{alg:get_index:basic}). The numerical value of the selection criteria for each $\mathit{hp}_\mathit{idx}$ hyper-parameter, henceforth denoted as `contribution', can be expressed as in Eq.\ref{eq:contribution:basic}:

\begin{equation}
\mathit{contribution}_\mathit{idx} = \mathit{q\_eval}_\mathit{idx} - \mathit{q\_best}
\label{eq:contribution:basic}
\end{equation}

\noindent
This results in $\mathit{contribution}_\mathit{idx} \in \left[ -1, 1 \right]$, with higher values being desirable.

However, a deep-learning algorithm can have a set of hyper-parameters, usually related to the network structure, increase (or decrease) in which incurs a computational cost. Using the above contribution formula could potentially cause a huge increase in the deep-learning algorithm hardware requirements or calculation time for a very small gain in terms of quality score. In such cases, an alternative method of contribution calculation may be employed (Eq.~\ref{eq:contribution:cost}).

\begin{equation}
\mathit{contribution}_\mathit{idx} = \left( \mathit{q\_eval}_\mathit{idx} - \mathit{q\_best} \right) \left( 1 - \theta_\mathit{idx}(\mathit{hp\_eval}_\mathit{idx}) \right)
\label{eq:contribution:cost}
\end{equation}

\noindent
$\theta_\mathit{idx}(\mathit{hp\_eval}_\mathit{idx})$ is a cost function, which output depends on the computational cost related to evaluated hyper-parameter value (Eq.~\ref{eq:theta}).

\begin{equation}
\theta_\mathit{idx}: \mathit{hp\_eval}_\mathit{idx} \rightarrow \mathit{cost} \land \mathit{cost} \in \left[ 0, 1 \right]
\label{eq:theta}
\end{equation}

Exact mapping done by the cost function can depend on factors such as observed deep-learning algorithm, hyper-parameter being adjusted, hardware being used etc. When $\theta_\mathit{idx}$ returns $0$, Eq.~\ref{eq:contribution:cost} is equivalent to Eq.~\ref{eq:contribution:basic}. As such, Alg.~\ref{alg:get_index:basic} can be replaced with Alg.~\ref{alg:get_index:cost} for all types of hyper-parameters, with appropriate $\theta_\mathit{idx}$ definitions.

\begin{algorithm}
\caption{Modified updated hyper-parameter selection}
\label{alg:get_index:cost}
\begin{algorithmic}[1]
\REQUIRE $\mathit{q\_eval}, \mathit{q\_best}, \mathit{hp\_eval}$
\ENSURE $\mathit{idx}$
\FOR{$i=0$ \TO $n-1$}
    \STATE $\mathit{contribution}_\mathit{i} = ( \mathit{q\_eval}_\mathit{i} - \mathit{q\_best} ) ( 1 - \theta_\mathit{i}(\mathit{hp\_eval}_\mathit{i}))$
\ENDFOR

\STATE $\mathit{idx} \leftarrow 0$ 
\FOR{$i=1$ \TO $n-1$}
    \IF{$\mathit{contribution}_{\mathit{idx}} < \mathit{contribution}_i$}
        \STATE $\mathit{idx} \leftarrow i$
    \ENDIF
\ENDFOR
\end{algorithmic}
\end{algorithm}
\section{Conclusions and Future Work}
\label{section:conclusions}

This paper introduces a concept of a new method to be used in a demanding and important task of deep learning algorithms hyper-parameters adjusting. The proposed method is based on an external agent denoted as `observer', which learns about an algorithm efficiency in terms of chosen quality score. This allows to model a performance of the algorithm with respect to its hyper-parameters. The author also proposes a method for incorporation of hardware resources consumption in the process of adjusting hyper-parameters. As a future work the author is going to implement the described method and conduct a series of experiments in order to compare its efficiency with other methods \cite{talathi2015hyper, young2015optimizing}.

\nocite{talathi2015hyper}

\bibliographystyle{elsarticle-num}
\bibliography{bibliography,authors-published,authors-unpublished}

\end{document}